\documentclass[sigconf]{acmart}
\AtBeginDocument{%
  }
\setcopyright{acmlicensed}
\copyrightyear{2025}
\acmYear{2025}
\acmDOI{XXXXXXX.XXXXXXX}

\usepackage{amsmath}
\usepackage{enumitem, booktabs}
\usepackage{graphicx}
\usepackage{hyperref}
\usepackage{url}
\usepackage{multirow}
\usepackage[table]{xcolor}
\usepackage{pifont}
\usepackage[utf8]{inputenc}
\usepackage[skip=8pt plus 2pt]{caption}

\newcommand{\Chatbot}{\textit{VayuChat}}
\newcommand{\chem}[2]{#1$_{#2}$}

\usepackage{etoolbox}
\newtoggle{CommentsMode}
\newcommand{\va}[1]{\iftoggle{CommentsMode}{{\color{green}\smaller \textbf{Vedant:} #1}}{}}

\acmConference[CODS 2025]{13th International Conference on Data Science}{December 17--20, 2025}{IISER, Pune, India}

\begin{document}

\title{\Chatbot{}: An LLM-Powered Conversational Interface for Air Quality Data Analytics}

\author{Vedant Acharya}
\authornote{Both authors contributed equally to this research.}
\email{vedant.acharya@iitgn.ac.in}
\affiliation{%
  \institution{Indian Institute of Technology Gandhinagar}
  \city{Gandhinagar}
  \state{Gujarat}
  \country{India}
}

\author{Abhay Pisharodi}
\authornotemark[1]
\email{abhay.pisharodi@iitgn.ac.in}
\affiliation{%
  \institution{Indian Institute of Technology Gandhinagar}
  \city{Gandhinagar}
  \state{Gujarat}
  \country{India}
}

\author{Rishabh Mondal}
\email{rishabh.mondal@iitgn.ac.in}
\affiliation{%
  \institution{Indian Institute of Technology Gandhinagar}
  \city{Gandhinagar}
  \state{Gujarat}
  \country{India}
}

\author{Mohammad Rafiuddin}
\email{mohammad.rafiuddin@ceew.in}
\affiliation{%
  \institution{Council on Energy, Environment and Water}
  \city{New Delhi}
  \state{Delhi}
  \country{India}
}

\author{Nipun Batra}
\email{nipun.batra@iitgn.ac.in}
\affiliation{%
  \institution{Indian Institute of Technology Gandhinagar}
  \city{Gandhinagar}
  \state{Gujarat}
  \country{India}
}

\renewcommand{\shortauthors}{Acharya et al.}

\begin{abstract}
Air pollution causes about 1.6 million premature deaths each year in India, yet decision makers struggle to turn dispersed data into decisions. Existing tools require expertise and provide static dashboards, leaving key policy questions unresolved. We present \textit{\Chatbot{}}, a conversational system that answers natural language questions on air quality, meteorology, and policy programs, and responds with both executable Python code and interactive visualizations. \Chatbot{} integrates data from Central Pollution Control Board (CPCB) monitoring stations, state-level demographics, and National Clean Air Programme (NCAP) funding records into a unified interface powered by large language models. Our live demonstration will show how users can perform complex environmental analytics through simple conversations, making data science accessible to policymakers, researchers, and citizens. The platform is publicly deployed at \url{https://huggingface.co/spaces/SustainabilityLabIITGN/VayuChat}. For further information check out video uploaded on \url{https://www.youtube.com/watch?v=d6rklL05cs4}.
\end{abstract}

\begin{CCSXML}
<ccs2012>
 <concept>
  <concept_id>10010147.10010178.10010187</concept_id>
  <concept_desc>Computing methodologies~Machine learning</concept_desc>
  <concept_significance>500</concept_significance>
 </concept>
 <concept>
  <concept_id>10010147.10010257.10010293.10010294</concept_id>
  <concept_desc>Computing methodologies~Natural language processing</concept_desc>
  <concept_significance>300</concept_significance>
 </concept>
</ccs2012>
</CCSXML>

\ccsdesc[500]{Computing methodologies~Machine learning}
\ccsdesc[300]{Computing methodologies~Natural language processing}
\end{CCSXML}
\keywords{Air Quality Analytics, Large Language Models, Conversational AI, Code Generation, Environmental Data}

\maketitle

\section{Introduction}
Air pollution in India causes over 1.6 million premature deaths annually and reduces life expectancy by more than five years due to \chem{PM}{2.5} exposure \cite{EPIC2023AQLIIndiaFactSheet,pandey2021health}.
The Central Pollution Control Board (CPCB)\footnote{A statutory body under the Government of India responsible for monitoring and regulating water and air pollution, operating over 500 stations under the National Air Monitoring Programme (NAMP).} continuously collects nationwide measurements of various pollutants \cite{cpcb_overview}. However, converting these raw datasets into timely, policy-relevant insights remains a challenge \cite{clarity_open_access_2022}. Even simple queries such as ``How did \chem{PM}{2.5} \va{change} vary last year in Delhi?'' remain hard to answer. In contrast, policy questions like ``Which cities reduced \chem{PM}{2.5} most relative to NCAP\footnote{Launched in 2019, NCAP allocated Rs.~9,650 crore to 131 non-attainment cities through FY 2019--24.} funding?'' require integrating pollution, funding, and population data. Existing tools often require expertise \cite{openair}, offer only limited visualizations \cite{naqi}, or handle only narrow tasks, leaving stakeholders dependent on specialists. Large language models (LLMs) can map natural-language queries to executable analyses~\cite{brown2020language, ouyang2022training}, yet their capacity for domain-specific, multi-dataset air quality analytics remains untested. Recent advances in instruction following~\cite{ouyang2022training} and zero-shot tabular reasoning~\cite{wei2023zero} show promise, but rigorous domain evaluation is needed for decision support.
\begin{figure*}[t]
    \centering
    \includegraphics[width=0.99\textwidth,clip,trim=0 30 0 0]{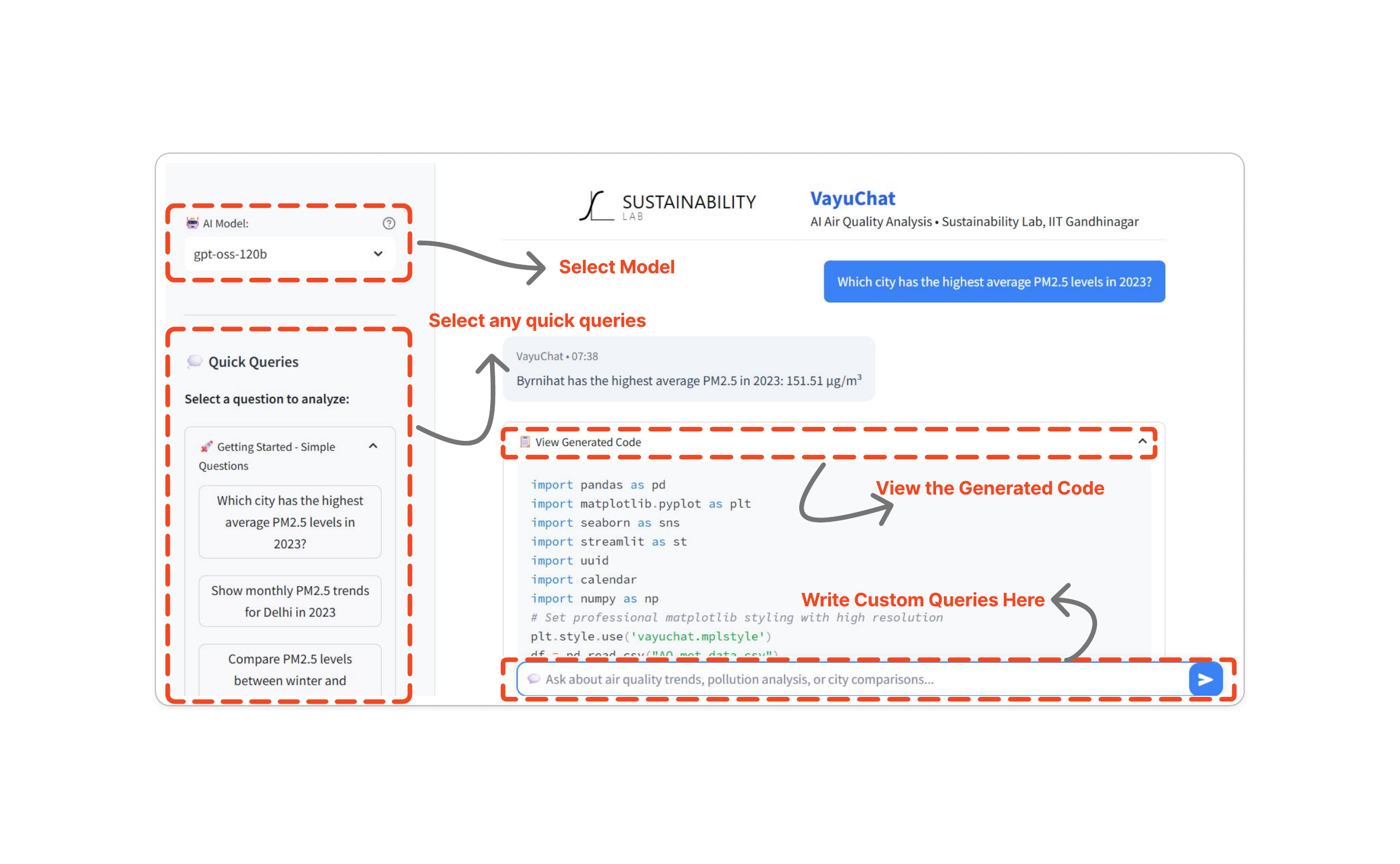}
    \vspace{-4em}
    \caption{\normalfont VayuChat Interface with key features: (1) \textbf{Select Model} - AI model dropdown (gpt-oss-120b), (2) \textbf{Quick Queries} - pre-defined air quality questions (e.g., PM2.5 comparisons, trends), (3) \textbf{Generated Code} - see generated Python code for analysis/visualization, and (4) \textbf{Custom Queries} - natural language input. Example query shows the highest \chem{PM}{2.5} level in 2023, with Byrnihat recording 151.51 $\mu$g/m\textsuperscript{3}.}
    \label{fig:vayuchat_interface}
\end{figure*}
We present \Chatbot{}, evaluated on our benchmark \textit{VayuBench} (whose details are beyond the scope of this paper), a system designed to fill this gap by offering a conversational interface that allows users to ask questions in natural language and receive both visual results and executable Python code. The primary motivation for using the code generation capabilities of LLMs in \Chatbot{} is to reduce hallucinations. Instead of giving the LLM the raw table, which can lead to hallucinations, we provide the dataset schema in the system prompt and have it generate Python code from the user’s query, ensuring reliable and transparent answers. A user can ask questions like ``Which cities had the worst air quality in winter 2023?'' or ``Which meteorological factor has the strongest correlation with the reduction of \chem{PM}{2.5} levels?'', and \Chatbot{} will return immediate, data-driven answers, accompanied by the underlying code. \Chatbot{} integrates three complementary datasets: (i) continuous air quality and meteorological observations from the Central Pollution Control Board (CPCB), (ii) population and area at the state level, and (iii) funding allocations under the National Clean Air Programme (NCAP). The conversational layer is powered by multiple state-of-the-art large language models, enabling comparative evaluation of alternative AI approaches for analytical tasks. This design facilitates rigorous assessment of model capabilities while lowering the barrier for policymakers, researchers, and citizens to perform complex analyses on air-quality datasets.

\section{System Description}
In this section, we describe the user interface of \Chatbot{} and its backend functionality. 
\subsection{User Interface}
As shown in Figure~\ref{fig:vayuchat_interface}, \Chatbot{} provides a browser-based interface where users can either select from quick query categories or enter their own custom air-quality related queries. Users can also choose the models they wish to use. Once a query is submitted, \Chatbot{} returns both the visual or textual answer and the corresponding code that generated the result.
\subsection{Backend}
\begin{figure}[h!]
  \centering
  \includegraphics[width=0.36\textwidth,keepaspectratio]{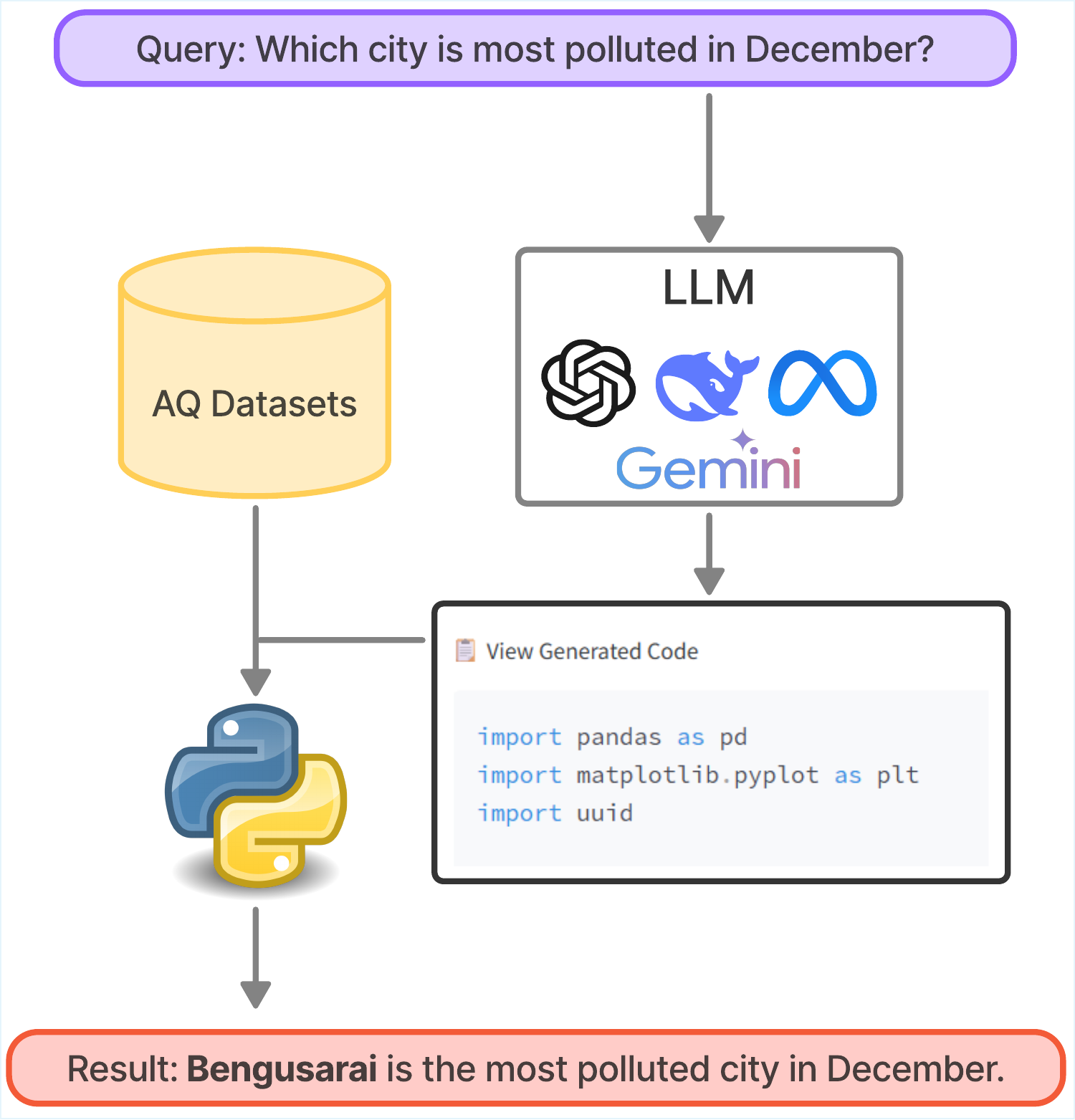}
  \Description{Flow diagram showing the workflow of the chatbot system.}
\caption{\normalfont Flow diagram of \Chatbot{} workflow: the user submits a query, the selected LLM generates code, which is executed in Python with the relevant dataset, and the output is presented.}

  \label{fig:flow_diagram}
\end{figure}

As shown in figure~\ref {fig:flow_diagram}, the backend of \Chatbot{} is responsible for processing user queries using the system prompt, invoking the appropriate model for response generation, and executing the code produced by the LLMs within a sandboxed environment. 

\subsubsection{Datasets and System Prompt}
In the system prompt, we provided descriptions of the datasets used. Below are the three datasets that we used in the system prompt:

\noindent\textbf{CPCB Data:}\hspace{1.0em} This dataset contains daily measurements of pollutants and weather variables for each CPCB monitoring station from 2017 to 2024~\cite{cpcb_aqi_repository}. Additionally, it includes station metadata, such as city, state, and the station ID assigned by CPCB. The detailed descriptions of the measured environmental variables, along with their units, are provided below: 
    \textit{Pollutants:} \chem{PM}{2.5} ($\mu$g/m\textsuperscript{3}), \chem{PM}{10} ($\mu$g/m\textsuperscript{3}), NO ($\mu$g/m\textsuperscript{3}), \chem{NO}{2} ($\mu$g/m\textsuperscript{3}), \chem{NO}{x} (ppb), \chem{NH}{3} ($\mu$g/m\textsuperscript{3}), \chem{SO}{2} ($\mu$g/m\textsuperscript{3}), CO (mg/m\textsuperscript{3}), Ozone ($\mu$g/m\textsuperscript{3}). \textit{Weather variables:}Air Temperature (AT, $^{\circ}$C), Relative Humidity (RH, \%), Wind Speed (WS, m/s), Wind Direction (WD, deg), Rainfall (RF, mm), Total Rainfall (TOT-RF, mm), Solar Radiation (SR, W/m\textsuperscript{2}), Barometric Pressure (BP, mmHg), and Vertical Wind Speed (VWS, m/s). 
    
\noindent\textbf{State population data:}\hspace{1.0em} This dataset provides information on the population and area (km\textsuperscript{2}) of 31 Indian regions according to the census data from 2011. It also includes a boolean column, \texttt{isUnionTerritory}, which indicates whether a region is a Union Territory.

\noindent\textbf{NCAP Funding:} \hspace{1.0em}This dataset provides financial year–wise records of funding released across Indian cities under the National Clean Air Program (NCAP) from 2019 to 2022. It includes columns for the total funds released and their utilization status as of June 2022.

\subsubsection{Models}
\Chatbot{} integrates a range of state-of-the-art models, including GPT-OSS 20B~\cite{openai2025gptoss120bgptoss20bmodel} and 120B~\cite{openai2025gptoss120bgptoss20bmodel}, Qwen3-32B~\cite{qwen3technicalreport}, Llama variants, DeepSeek-R1~\cite{deepseekai2025deepseekr1incentivizingreasoningcapability}, and the latest Gemini models. The Gemini models are accessed through the Gemini API, while the other open-source models are accessed via the Groq Cloud API~\cite{GroqDocs}.  
\subsubsection{Response Generation}
When a user submits a query, it is combined with the system prompt, which includes instructions that the dataframes are already loaded and passed to the selected model. The model then generates a Python code snippet based on the query. The generated code is stored and made accessible to the user, ensuring transparency and reproducibility of the results.

\subsubsection{Code Execution}
\Chatbot{} executes the code generated by the LLM in a secure, sandboxed environment to prevent any potential system harm. The execution results are captured and integrated into the response. Depending on the nature of the query and the output produced, the results are presented to the user in different formats such as text, structured tables, or visualizations (e.g., plots and charts). 

\subsubsection{Plot Generation}
The \Chatbot{} system automatically generates visualizations in response to relevant analytical queries. Figure~\ref{fig:wind_pm25} and Figure~\ref{fig:five_year_trend} demonstrate examples of plot-based responses, where the system produces graphical outputs corresponding to user prompts requesting data visualization.

\subsubsection{Capabilities of VayuChat}

VayuChat analyzes comprehensive air-quality and meteorological data (2017–2024), including CPCB pollutant measurements, weather parameters, and state population and area information. Its capabilities can be grouped as follows:

\begin{itemize}
    \item[\checkmark] \textit{Direct queries:} ``Which city had the highest \chem{PM}{2.5} in 2023?',; ``Show \chem{SO}{2} levels for Delhi.''
    \item[\checkmark] \textit{Plot generation:} ``Plot \chem{PM}{2.5} trends for Mumbai'', ``Compare ozone levels between Punjab and Gujarat.''
    \item[\checkmark] \textit{Analytical queries:} ``Analyze wind speed and \chem{PM}{2.5} correlation''; ``Evaluate NCAP impact on air quality.''
\end{itemize}

The system generates both Python code and visualizations through code-generation LLMs, ensuring transparency and reproducibility.

\section{VayuChat Case Study: Delhi Air Quality Analysis}
% \nb{We evaluated VayuChat in two wa...: i) LLM perfornmance ... VayuBench .. (outside scope) where we found ....; ii) working with experts and users...}\ap{Added}
% \rafi{sample comment}
We worked with air quality analysts to investigate the reasons behind Delhi's severe pollution spike in December 2024 using our platform, \Chatbot{}. In this section, we present the insights generated through \Chatbot{} by asking targeted questions that break down the causes of the crisis. Our analysis uses CPCB data from 37 monitoring stations across Delhi, together with meteorological observations, to reveal significant trends.
% We evaluated VayuChat in two ways: i) LLM performance evaluation using VayuBench (outside scope of this paper) where we found [brief finding]; ii) working with experts and users in real-world analytical scenarios. Our investigation proceeded through a systematic series of interconnected questions that progressively deepened the analysis: \textbf{1. Initial identification:} "Which days in Delhi are the most polluted in December 2024?" - to pinpoint the crisis period. \textbf{2. Temporal pattern analysis:} "Use a time series plot to compare the pollution levels and wind speed during Delhi's most polluted week in December 2024 to the previous and after 15 days?" - to understand short-term dynamics. \textbf{3. Historical contextualization:} "Plot and compare the pollution levels and wind speed in Delhi during a polluted week in December 2024 with the data from the previous five years" - to assess crisis severity relative to historical patterns. \textbf{4. Multi-pollutant correlation:} "Analyze the correlation between CO, NO2, and PM2.5 levels in Delhi for the month of December from 2017 onwards" - to understand pollutant interaction patterns.
We evaluated VayuChat in two complementary ways:  
(i) by assessing LLM performance using \textit{VayuBench} (details outside the scope of this paper), and  (ii) by collaborating with experts and end users in real-world analytical scenarios. Our investigation proceeded through a systematic series of interconnected questions that progressively deepened the analysis:

\noindent \textbf{Initial Identification:} ``Which days in Delhi were the most polluted in December 2024?'' This pinpoints the crisis period.  

\noindent \textbf{Temporal Pattern Analysis:} ``Use a time series plot to compare pollution levels and wind speed during Delhi’s most polluted week in December 2024 to the preceding and following 15 days.'' This examines short-term dynamics.  

\noindent \textbf{Historical Contextualization:} ``Plot and compare pollution levels and wind speed in Delhi during a polluted week in December 2024 with the data from the previous five years.'' Assesses crisis severity relative to historical patterns.  

\noindent \textbf{Multi-Pollutant Correlation:} ``Analyze the correlation between CO, NO\textsubscript{2}, and PM\textsubscript{2.5} levels in Delhi for December from 2017 onwards.'' Investigates pollutant interaction patterns.  

This sequential questioning approach allowed analysts to build understanding from crisis identification to mechanistic insights. In future, we envision the system to automatically reason out such sub-questions to provide complete investigation workflows. A striking finding is the strong inverse relationship between wind speed and \chem{PM}{2.5}: even a 0.6 m/s drop trapped pollutants, shifting air quality from ``Very Poor'' to ``Severe'' within hours. These results show how subtle weather changes can turn Delhi's winter air into a hazardous blanket for 30 million residents, offering insights for both emergency planning and long-term policy.

\subsection{Most Polluted Dates in December 2024 Delhi}

\textbf{User Prompt:}\hspace{1.0em} Which days in Delhi are the most polluted in December 2024?
% \textcolor{blue}{Abhay for consistency purpose better to give SS}

\noindent\textbf{\Chatbot{} Response:}\hspace{1.0em}Based on our analysis of December 2024 data, the most severely polluted dates in Delhi are presented in \textbf{Table \ref{tab:polluted_dates}}.

\begin{table}[h]
\centering
\begin{tabular}{@{}cr@{}}
\toprule
\textbf{Date} & \textbf{\chem{PM}{2.5} ($\mu$g/m\textsuperscript{3})} \\
\midrule
18-12-2024 & 344.59 \\
19-12-2024 & 341.46 \\
17-12-2024 & 330.25 \\
20-12-2024 & 291.46 \\
22-12-2024 & 285.98 \\
\bottomrule
\end{tabular}
\caption{\normalfont (\textbf{Generated by VayuChat}) Most Polluted Dates in December 2024 Delhi}
\label{tab:polluted_dates}
\end{table}

\subsection{Wind Speed Analysis: Most Polluted Week vs. Surrounding Periods}

\textbf{User Prompt:}\hspace{1.0em} Use a time series plot to compare the pollution levels and wind speed during Delhi's most polluted week in December 2024 to the previous and after 15 days?

\begin{figure}[h]
\centering
\includegraphics[width=\columnwidth]{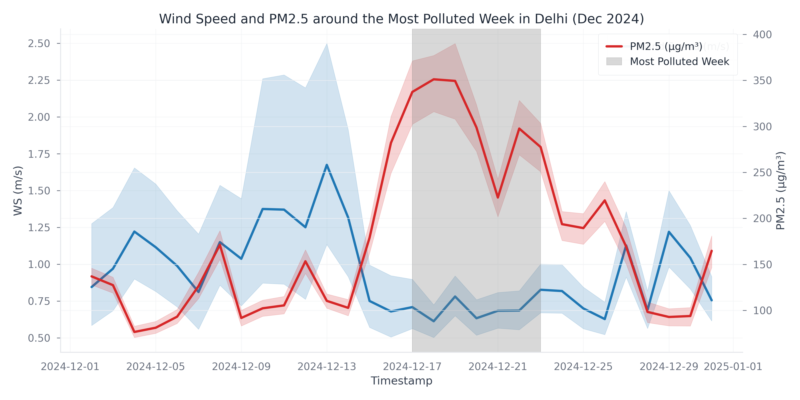}
\caption{(\textbf{Generated by VayuChat}) \normalfont Wind Speed vs \chem{PM}{2.5} Concentration - December 2024 Critical Week}
\label{fig:wind_pm25}   
\end{figure}

\noindent\textbf{\Chatbot{} Response:}\hspace{1.0em}The most polluted week was identified in December 2024 (see Fig.~\ref{fig:wind_pm25}).  
% \noindent\textbf{Key Observation:} Clear inverse relationship between wind speed and pollution levels - when wind speeds dropped below 1.0 m/s, \chem{PM}{2.5} levels consistently exceeded 300 $\mu$g/m\textsuperscript{3}.

\vspace{-1.5em} % adjust as needed
\noindent\textbf{Key Observation:} \hspace{1.0em} A clear inverse relationship exists between wind speed and \chem{PM}{2.5}. When wind speeds dropped below 1.0~m/s, \chem{PM}{2.5} levels exceeded 300~$\mu$g/m\textsuperscript{3}, as stagnant conditions trapped pollutants near the surface. Even small decreases in wind speed rapidly pushed air quality from ``Very Poor'' to ``Severe,'', showing the sensitivity of Delhi’s winter pollution to meteorological variations.
\subsection{Five-Year Comparison: December Pollution and Wind Speed Trends}

\textbf{User Prompt:} Plot and compare the pollution levels and wind speed in Delhi during a polluted week in December 2024 with the data from the previous five years.

\begin{figure}[h]
\centering
\includegraphics[width=0.90\columnwidth]{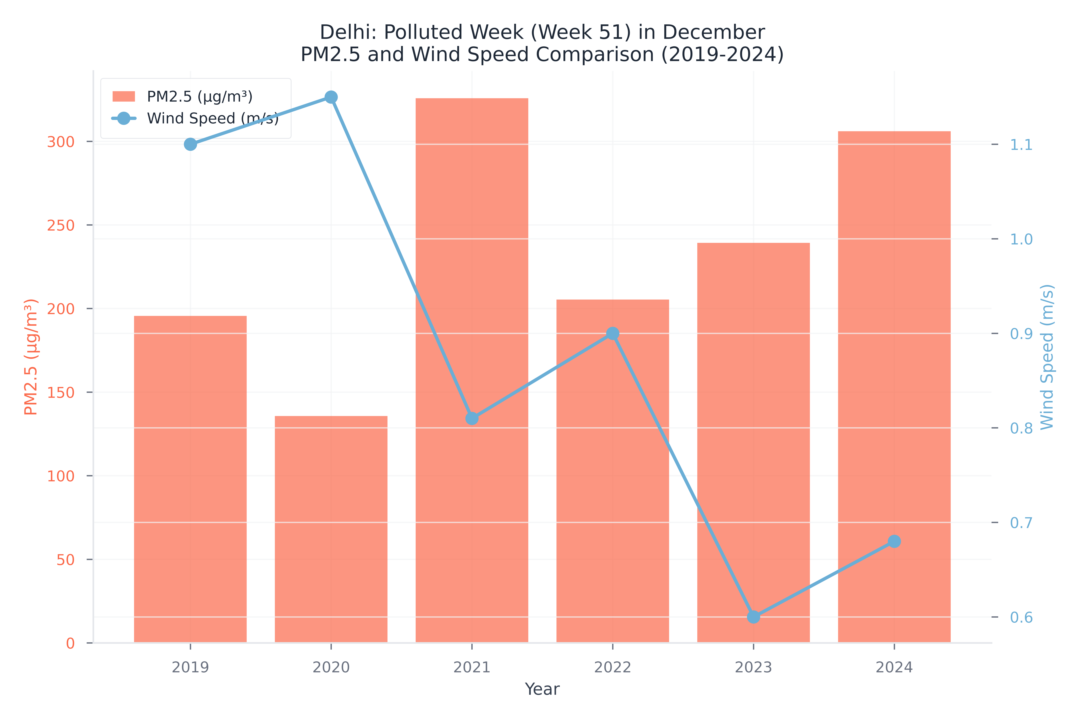}
\caption{(\textbf{Generated by VayuChat}) \normalfont Five-Year December \chem{PM}{2.5} and Wind Speed Trends (2019-2024)}
\label{fig:five_year_trend}
\end{figure}

\noindent\textbf{\Chatbot{} Response:}\hspace{1.0em} Comparison of December 2024 with previous years (2019–2023), as shown in Fig.~\ref{fig:five_year_trend}.

\vspace{-1.2em}
\noindent\textbf{Key Observation:}\hspace{1.0em} In 2024, wind speeds showed marginal improvement compared to previous years.  \va{A strong correlation was evident, with years of higher wind speeds (2019 and 2020) corresponding to relatively lower \chem{PM}{2.5} concentrations.} While we observe a strong negative correlation between \chem{PM}{2.5}  and ind speed during 2019 and 2020, a more comprehensive analysis requires considering both temperature and wind speed. The lowest wind speeds (0.6 m/s) were recorded in 2023, while 2021 had the highest PM\textsubscript{2.5} levels at 325~$\mu$g/m\textsuperscript{3}.
% \noindent\textbf{Key Observation:}
% \begin{itemize}
% \item 2024 showed marginal improvement compared to previous years in wind speed.
% \item Strong correlation: Years with higher wind speeds (2019, 2020) had relatively lower \chem{PM}{2.5} concentrations
% \item 2023 was the worst year with lowest wind speeds (0.6 m/s) and 2021 was the highest \chem{PM}{2.5} (325 $\mu$g/m\textsuperscript{3}).
% \end{itemize}
\subsection{Correlation Analysis: CO, NO2, and PM2.5 (December 2017-2024)}

\textbf{User Prompt:}\hspace{1.0em} Analyze the correlation between CO, NO\textsubscript{2}, and PM\textsubscript{2.5} levels in Delhi during December from 2017 onwards, as presented in Table~\ref{tab:correlation}.

\begin{table}[h]
\centering
\setlength{\tabcolsep}{4pt} % default is 6pt; reduce to tighten

\begin{tabular}{@{}lrrr@{}}
\toprule
\textbf{Pollutants} & \textbf{CO (mg/m\textsuperscript{3})} & \textbf{\chem{NO}{2} ($\mu$g/m\textsuperscript{3})} & \textbf{\chem{PM}{2.5} ($\mu$g/m\textsuperscript{3})} \\
\midrule
CO (mg/m\textsuperscript{3}) & 1 & 0.3 & 0.47 \\
\chem{NO}{2} ($\mu$g/m\textsuperscript{3}) & 0.3 & 1 & 0.34 \\
\chem{PM}{2.5} ($\mu$g/m\textsuperscript{3}) & 0.47 & 0.34 & 1 \\
\bottomrule
\end{tabular}%
\caption{(\textbf{Generated by VayuChat}) \normalfont Correlation Matrix: CO, \chem{NO}{2}, and \chem{PM}{2.5} (December 2017--2024).}
\label{tab:correlation}
\end{table}
\vspace{-2.7em} % adjust as needed

\noindent\textbf{Key Observation:}\hspace{1.0em} The analysis shows strong inter-correlation among all three pollutants in December, with \chem{PM}{2.5} and CO most related (r = 0.47). This suggests common sources such as vehicles, stubble burning, and industrial emissions drive synchronized pollution episodes.
\section{Potential Applications}
\Chatbot{} serves diverse communities in three ways: (1) \textit{Government and policy}, enabling officials to evaluate programs, compare outcomes, and prioritize interventions (e.g., “Which states improved most after NCAP funding?”); (2) \textit{Journalism and public engagement}, helping media create data-driven stories and citizens interpret local air trends without technical expertise; and (3) \textit{Education and community monitoring}, serving as both a teaching resource in data science and environmental studies and a tool for organizations to track conditions, identify sources, and advocate interventions.

% \section{Potential Applications}
% \Chatbot{} addresses real-world needs across multiple user communities: 
% \begin{itemize}
%     \item \textbf{Government and Policy:} Officials can quickly assess policy effectiveness, compare regional performance, and identify intervention priorities. For example: \textit{"Which states improved most after receiving NCAP funding?"} or \textit{"What seasonal patterns should inform our pollution control strategies?"} 
%     \item \textbf{Scientific Research:} Researchers can rapidly prototype analyses, explore hypotheses, and generate publication-ready visualizations. The transparent code generation enables reproducible research workflows and collaborative analysis. \item \textbf{Journalism and Public Engagement:} Media professionals can create data-driven stories about environmental issues, while citizens can understand air quality trends in their regions without technical expertise.
%     \item \textbf{Education and Training:} The platform serves as a teaching tool for data science, environmental studies, and policy analysis, demonstrating best practices through real-world examples.
%     \item \textbf{Community Monitoring:} Local organizations and activists can track environmental conditions, identify pollution sources, and advocate for targeted interventions based on data evidence.
% \end{itemize}
\section{Future Work}
In the future, we aim to enhance \Chatbot{} in four ways: (1) \textit{Live dataset integration}, incorporating real-time air quality streams via APIs; (2) \textit{Expanding datasets}, adding ERA5 reanalysis, satellite products, and land-use or emission inventories for richer spatial and temporal coverage; (3) \textit{Model finetuning}, tailoring LLMs for accurate code generation and domain-specific query interpretation; and (4) \textit{Automated reasoning workflows}, enabling decomposition of complex queries into sub-questions and multi-step analyses, as shown in our Delhi pollution case study.

\section{Conclusion}
\Chatbot{} advances accessible environmental analytics while maintaining scientific rigor, coupling conversational AI with integrated datasets to enable complex analyses without technical skills. Transparent code generation, multiple models support, and domain-specific design make it suitable for policy, research, public engagement, and education, demonstrating that natural language queries can yield reproducible analyses and actionable insights on complex air-quality datasets.

\bibliographystyle{ACM-Reference-Format}
\bibliography{references}

\end{document}